\documentclass{article}

\usepackage{arxiv}
\usepackage{graphicx}
\usepackage{subcaption}
\usepackage{cite}
\usepackage{amsmath}
\usepackage{amssymb}
\usepackage{mathtools}
\usepackage{stfloats}
\usepackage{lipsum}
\usepackage{comment}

\title{Modeling and Control of a Pneumatic Morphing Soft Quadrotor based on the SOFA Framework for Dynamic Soft Robotic Simulation}

\author{
  Fernando Labra Caso%
  \thanks{This work has been partially funded by the European Union’s Horizon Europe Research and Innovation Program, under the Grant Agreement No. 101119774 SPEAR.} \\
  Robotics \& AI\\
  Luleå University of Technology \\
  \texttt{fernando.labra.caso@ltu.se} \\
   \And
  Vidya Sumathy \\
  Robotics \& AI\\
  Luleå University of Technology \\
  \texttt{vidya.sumathy@ltu.se} \\
   \And
  Pasquale Ferrentino \\
  IMEC Center \\
  VUB Brussels University \\
  \texttt{pasquale.ferrentino@vub.be} \\
   \And
  Bram Vanderborght \\
  IMEC Center\\
  VUB Brussels University \\
  \texttt{bram.vanderborght@vub.be} \\
   \And
  Jakub Haluska \\
  Robotics \& AI\\
  Luleå University of Technology \\
  \texttt{jakub.haluska@ltu.se} \\
   \And
  George Nikolakopoulos \\
  Robotics \& AI\\
  Luleå University of Technology \\
  \texttt{george.nikolakopoulos@ltu.se} \\
}

\begin{document}

\maketitle

\begin{abstract}
This article presents a novel SOFA based finite element method for the soft body modeling and the corresponding dynamic simulation and control of a pneumatic morphing soft quadrotor. The proposed modeling preserves the physical interpretability and control structure of traditional quadrotor dynamics, while capturing the complex, time-varying behavior of pneumatically actuated soft arms. In SOFA, the soft pneumatically actuated arms are discretized as a tetrahedral mesh following an elastic material law that produces internal forces adequate to the real dynamic behavior of the body. Pneumatic actuation governed by both periodic and error-based control signals is applied within the internal cavities to analyze the morphing capability. Finally, a proportional-integral controller is proposed to study the controlled dynamic behavior and morphing capabilities of the pneumatic arm, wherein the pneumatic actuation to the soft arm is controlled to achieve the desired target position. The simulation results show the effectiveness of the proposed novel modeling framework and the related controller design.
\end{abstract}

\twocolumn

\section{Introduction}
\noindent Inspired from nature, soft UAV's are more compliant in design and have high thrust-to-weight ratio, but this comes at the cost of more complex dynamics and control challenges, compared to conventional UAV's~\cite{tanaka2022review, hammad2024landing}. One of the best characteristics of soft drones has been the ability to deform and navigate through confined spaces and the potential to withstand collisions, hence, they are used in applications like search and rescue~\cite{fabris2021soft}, full-body perching on pipelines and irregular surfaces~\cite{ruiz2022ieee, zheng2023metamorphic}, aerial physical interaction~\cite{bredenbeck2024embodying} and safe interactions like grasping ~\cite{cheung2024modular,xu2024soft}. 

Soft robots, in contrast to conventional rigid robots, are defined by the integration of compliant materials~\cite{xing2024morphing}, actuation modalities~\cite{chen2023bioinspired}, sophisticated modeling techniques~\cite{armanini2023soft, mengaldo2022concise}, and adaptive control strategies~\cite{della2023model,wang2022control}, often guided by bio-inspired principles~\cite{ahmed2022decade} to optimize flexibility, adaptability, and functional versatility~\cite{yasa2023overview}. In comparison to rigid robots, soft robots have more complex dynamics and henceforth understanding, modeling, predicting or replicating the soft behavior is extremely complex. A broad review of various approaches to soft robot modeling, which includes continuum models, geometric models, discrete material models and surrogate models, can be found in~\cite{armanini2023soft}. The work in~\cite{chen2024data} presented a review of different data-driven methods and control approaches for soft robots, while~\cite{della2023model} presented a survey the state-of-the-art and related challenges in model-based control of soft robots. 

SOFA (Simulation Open Framework Architecture) is a widely used software to simulate soft-robots. The works in~\cite{ferrentino2023finite, ferrentino2022quasi} presented a Finite Element Analysis (FEA) modeling of a soft actuator and a quasi-static FEA model of a multi-material soft pneumatic actuator in SOFA, respectively. Furthermore, a finite element model for the dynamics of concentric tube robots and their interaction with deformable environment using SOFA is presented in~\cite{zuo2022finite}. The work in~\cite{wehner2025open} presented an interface for simulating the model of magnetic soft robots in SOFA that allows users to define material properties, apply magnetic fields and observe resulting deformations in real time, while~\cite{mokhtar2025development} presented the design and development of a SOFA scene as a gym environment to train the model of a conceptual tripedal robot using the PPO reinforcement learning algorithm.

\begin{figure*}[!t]
    \centering    \includegraphics[width=\linewidth]{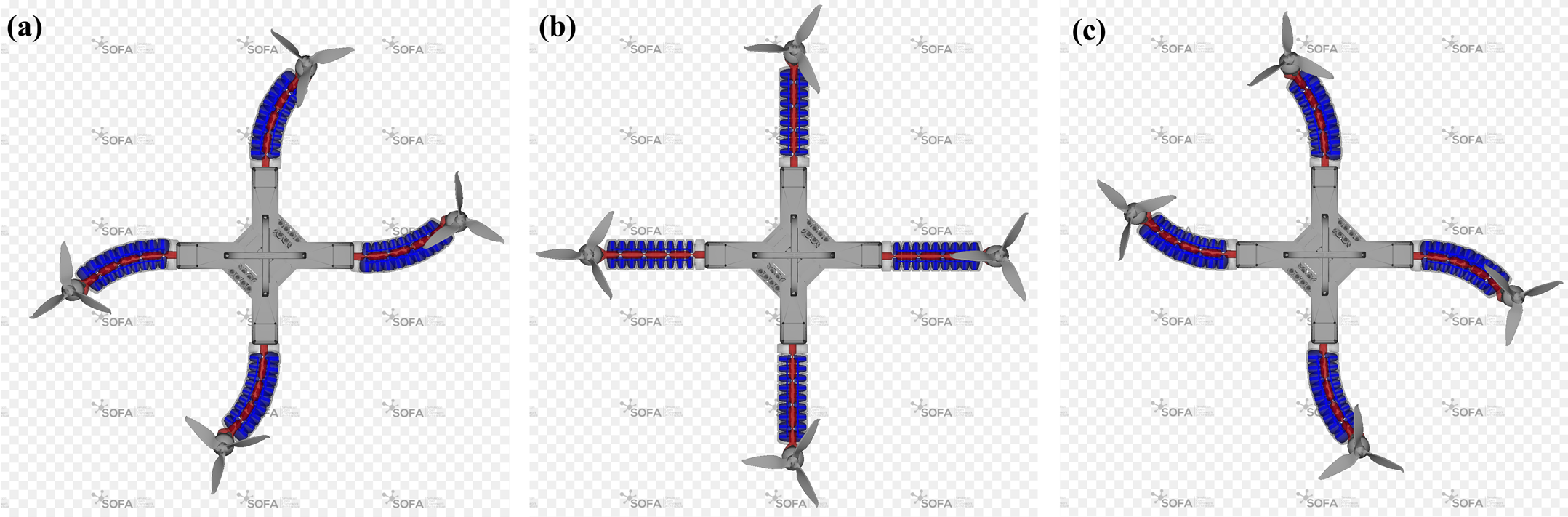}
    \caption{Visualization of the presented morphing pneumatic quadrotor design in SOFA at initial state (b) and during actuation of the lateral pressure cavities under two configurations (a), (c).} \vspace{-6mm}
    \label{fig:figure2}
\end{figure*}

In contrast to the presented state of the art, this research work presents a totally novel FEM-based soft body modeling and dynamic simulation approach for a Pneumatic Morphing Soft Quadrotor (PMSQ) using the SOFA framework, with an overview of this depicted in Figure~\ref{fig:figure2}, for both the unactuated and actuated configurations. The presented framework for the dynamic simulation could be utilized for the evaluation of the PMSQ's in-flight morphing capabilities using pneumatic control. As the quadrotor has a symmetric structure, and without loosing any generality or scientific contribution, in this article, only the one arm will be modeled and replicated along the edges of the simulated drone. Thus, the main contribution includes the novel finite-element modeling of a soft morphing aerial platform capturing coupled pressure–structure behavior in pneumatically actuated arms using SOFA. The second contribution is focusing on the establishment of a a fundamental closed-loop position controller for a single cavity pneumatic control and its validation through SOFA modeling under dynamic FEM simulations.

\section{Methodology}

\subsection{The SOFA Modeling Framework}

\begin{figure}[h!]
    \centering
    \includegraphics[width=\linewidth]{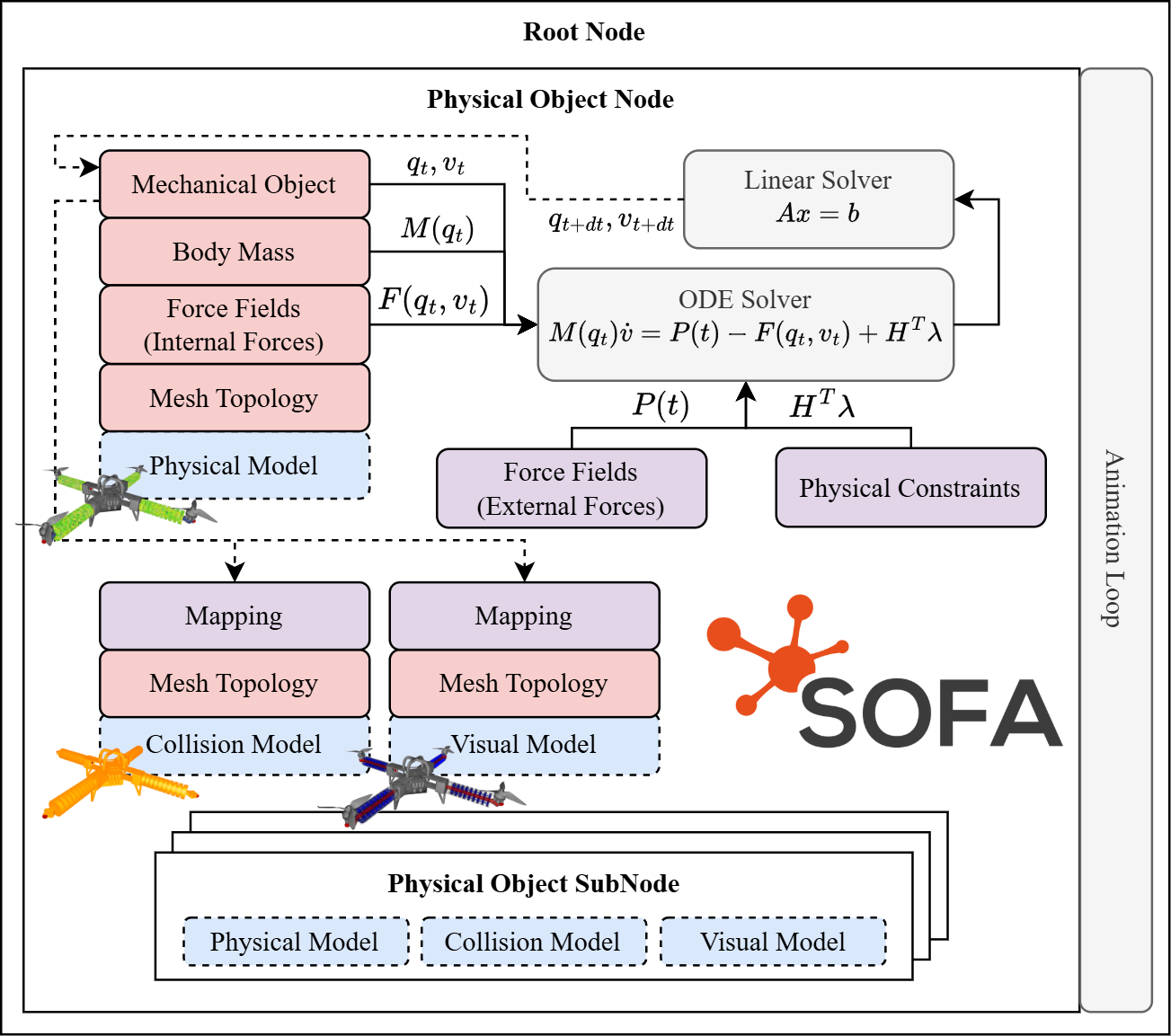}
    \caption{Overview of the components in a SOFA scene during soft FEM simulations. The process is governed by the time integration of every dynamic single body behavior. Different components defining forces, body masses, joints or position constraints allow realistic soft robotic representations.} \vspace{-3mm}
    \label{fig:figure8}
\end{figure}

\noindent A SOFA FEM scene can be understood as a modular, layered system in which a central mechanical state is progressively transformed by physics operators through ODE integration and linear solving, while Figure~\ref{fig:figure8} displays an overview of such soft-body scene. The deformable object is discretized over a mesh or topology structure, which defines the connectivity of elements used during the FEM simulation. The mechanical state stores the degrees of freedom (DOFs) of the system, typically positions $q \in \mathbb{R}^n$ and velocities $v \in \mathbb{R}^n$, according to the topology discretization. Force models and constraints rely on this topological information to determine how deformations propagate across the structure, while several physics operators are used to describe the overall object's behavior. Mass properties introduce inertia $M(q): \mathbb{R}^n \rightarrow M^{n \times n}$, force models compute both internal $F(q_t,v_t)$ (e.g. elastic responses), and external $P(t)$ (e.g. gravity or user-defined loads) contributions. Constraints $H^T\lambda$ further restrict the admissible motion of the system by imposing conditions, such as fixed regions, attachments, or contact interactions. In this formulation $H^T$ represents the constraint direction matrix and $\lambda \in  \mathbb{R}^n$ the constraint force intensities. Together, mass, forces, and constraints define the physical model at each simulation step.

The system advances through an Ordinary Differential Equation (ODE) solver, which determines how the mechanical state evolves. The system follows Newton's second law for modeling the dynamic behavior of a body, as in Eq. \ref{Eq1}:
\begin{equation}\label{Eq1}
\underbrace{M(q_t)}_{\text{Inertia Matrix}}\dot{v}=\underbrace{P(t)}_{\text{External Forces}}-\underbrace{F(q_t,v_t)}_{\text{Internal Forces}}+\underbrace{H^T\lambda}_{\text{Constraints}}    
\end{equation}

This work utilizes an Euler implicit solver where forces are considered based on the state information at the next time step $x(t+dt)$, being unknown at the current time step. Constraints define i.a. attachments between bodies, generating a set of unknown variables that need to be solved at each time step. A linear solver is then utilized to calculate such variables and compute the necessary increments or corrections to the DOFs, while the result dynamically updates the mechanical state consistently. In addition to the physical representation, the scene includes auxiliary geometric models dedicated to collision detection and visualization. These alternate geometries reduce the computational load by simplifying the meshes from tetrahedral to triangular or point-based topologies. The collision model enables efficient contact detection and response whereas the visual model allows higher-resolution renderings. These models do not independently define physical behavior; but follow the updates of the mechanical state. Consistency between the mechanical DOFs and these auxiliary representations is ensured through mappings, denoted as $p=J(q)$, with $J$ a non-linear function. Mappings propagate the motion of the mechanical state and transfer interaction effects, such as collision responses, back towards the DOFs. The interaction between layers is coordinated by an animation loop. At each $dt$, the loop enforces a sequence of operations: evaluation of forces and constraints from the current state, collision detection and response, invocation of the ODE solver and any required linear solve, update of the mechanical DOFs, and propagation of the updated states to mapped representations. This orchestration, presented in Figure~\ref{fig:figure8} ensures that all components interact in a consistent and deterministic order. 

\begin{figure*}[!t]
    \centering
    \includegraphics[width=\linewidth]{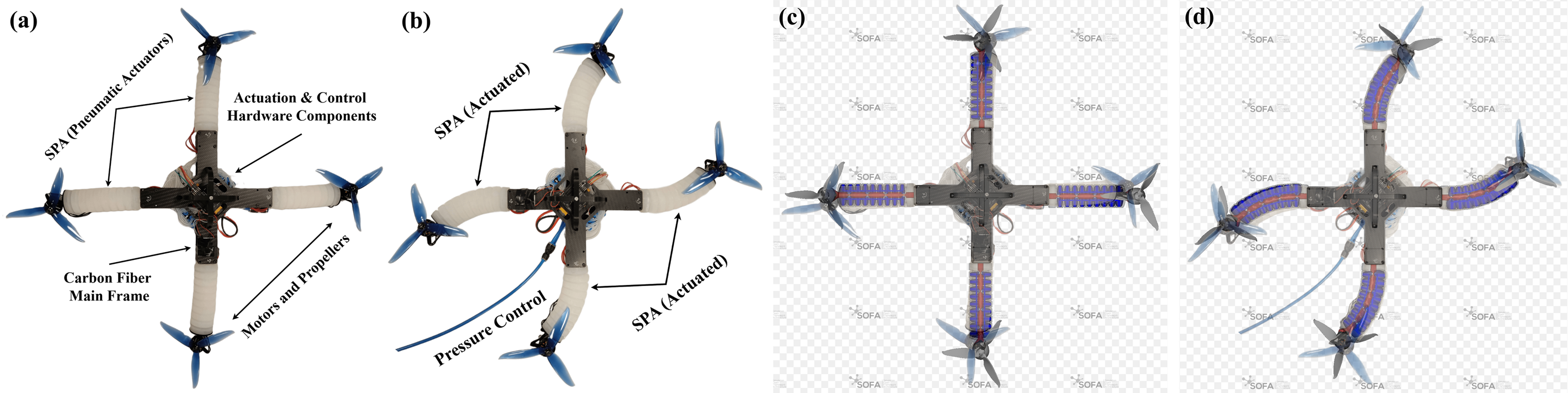}
    \caption{Soft Morphing Quadrotor at initial state (a) and during actuation (b). The Soft Pneumatic Arms connect to a carbon fiber frame containing the pressure actuation and motor control hardware components. The SOFA design overlay (c), (d) demonstrates the modeling capabilities of the SOFA framework for soft robotic validation under dynamic FEM simulations.} \vspace{-4mm}
    \label{fig:figure1}
\end{figure*} 

\begin{figure*}[b!]
    \centering
    \includegraphics[width=\linewidth]{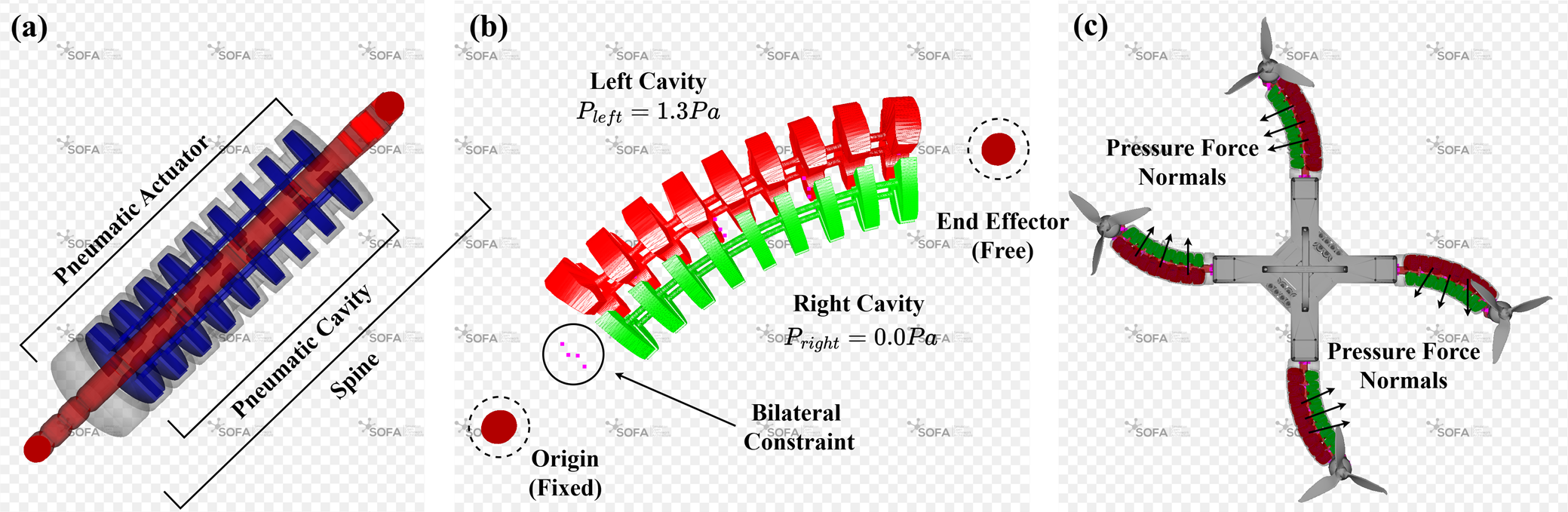}
    \caption{(a) The different parts of the UAV arm modelled in SOFA. The pneumatic actuator (white), the pneumatic cavity chambers (blue), and the rigid spine (red). (b) The spine and the pneumatic actuator are constrained together with a bilateral Lagrangian constraint. One edge of the spine is fixed in space while the distal edge is free to bend. (c) The cavities will bend the arm by differential pressure $\Delta P = P_{left} - P_{right}$, with resulting forces following the normals of the mesh.}
    \label{fig:figure3}
\end{figure*}

\subsection{Pneumatic Morphing Soft Quadrotor}
\noindent The pneumatic morphing soft quadrotor utilized in this research follows the initial established design presented in~\cite{Haluska2022med}. The platform consists of a central carbon-fiber frame supporting four pneumatically actuated soft arms, with motors and propellers mounted at their free ends, as visible in Figure~\ref{fig:figure1}(a). Alongside standard electronic components, including microcontrollers, transmitters and receivers, and electronic speed controllers; the pneumatic valves, driving the soft arms, are integrated into the PMSQ hardware. The complete hardware configuration includes a Pixracer R15 flight controller, four T-motor F40PRO IV kV2400 motors fitted with T5150 three-blade propellers, a T-motor P50A 6S 4IN1 ESC, and a ZIPPY Compact 1400 mAh 4S 65C LiPo battery. The total weight of the assembled system is approximately 1 kg.

The Soft Pneumatic Arms (SPA) are integrated into the UAV to enable morphing capabilities and have two main components: 1) A  semi-rigid inner spine, 3D-printed in nylon, which connects the propeller with the main drone frame; providing structural support against vertical loads, while still allowing motion in the horizontal plane. 2) A pneumatic actuator, molded in DragonSkin 30, and attached to the spine. Each actuator contains four air chambers with pneumatic input channels.

\subsection{SOFA Modeling Design}

\noindent The SOFA FEM simulator~\cite{faure2012sofa, ferrentino2023finite} is employed to model the mechanical behavior of the soft drone under dynamic conditions. The SOFA scene is composed of a rigid drone box, fixed in space with four arms, each  containing a pneumatic actuator, four pressure cavities and a structural spine.

The physical model of the pneumatic actuator is represented by a tetrahedral mesh composed of 19466 elements, as depicted in Figure \ref{fig:figure4}(a). The material properties of the DragonSkin 30 silicone used for the Soft Pneumatic Actuator (SPA) are modeled using the \textit{Stable Neohookean} hyperelastic constitutive law \cite{ferrentino2021fea}, with parameters $\mu = 0.24203$ [MPa] and $\lambda = 0$ [MPa] (incompressibility assumption). The internal cavities of the pneumatic chambers, shown in Figure  \ref{fig:figure3}(b), are meshed with 8056 triangular elements, and the internal pressure is applied through the SOFA component \textit{Surface Pressure Constraint}.

The physical model of the structural spine is defined using a separate tetrahedral mesh of 7648 elements. The Thermoplastic Polyurethane (TPU) composing the spine is modeled as a linear elastic material, characterized by a Young’s modulus $E = 75$ [MPa] and a Poisson’s ratio $\nu = 0.45$. The spine is connected to the SPA via a series of \textit{Bilateral Interaction Constraints}, labelled in Figure \ref{fig:figure3}(b), which ensure mechanical coupling between the two components.

Each component implements collision and visual models with triangular mesh topology. The mapping utilized between tetrahedral and triangular DOFs follows a barycentric scheme where several points from the physical model associate with singular points in the auxiliary geometrical models.

\section{Results}

\subsection{Periodic Pneumatic Actuation}
\begin{figure}[!h]
    \centering
    \includegraphics[width=\linewidth]{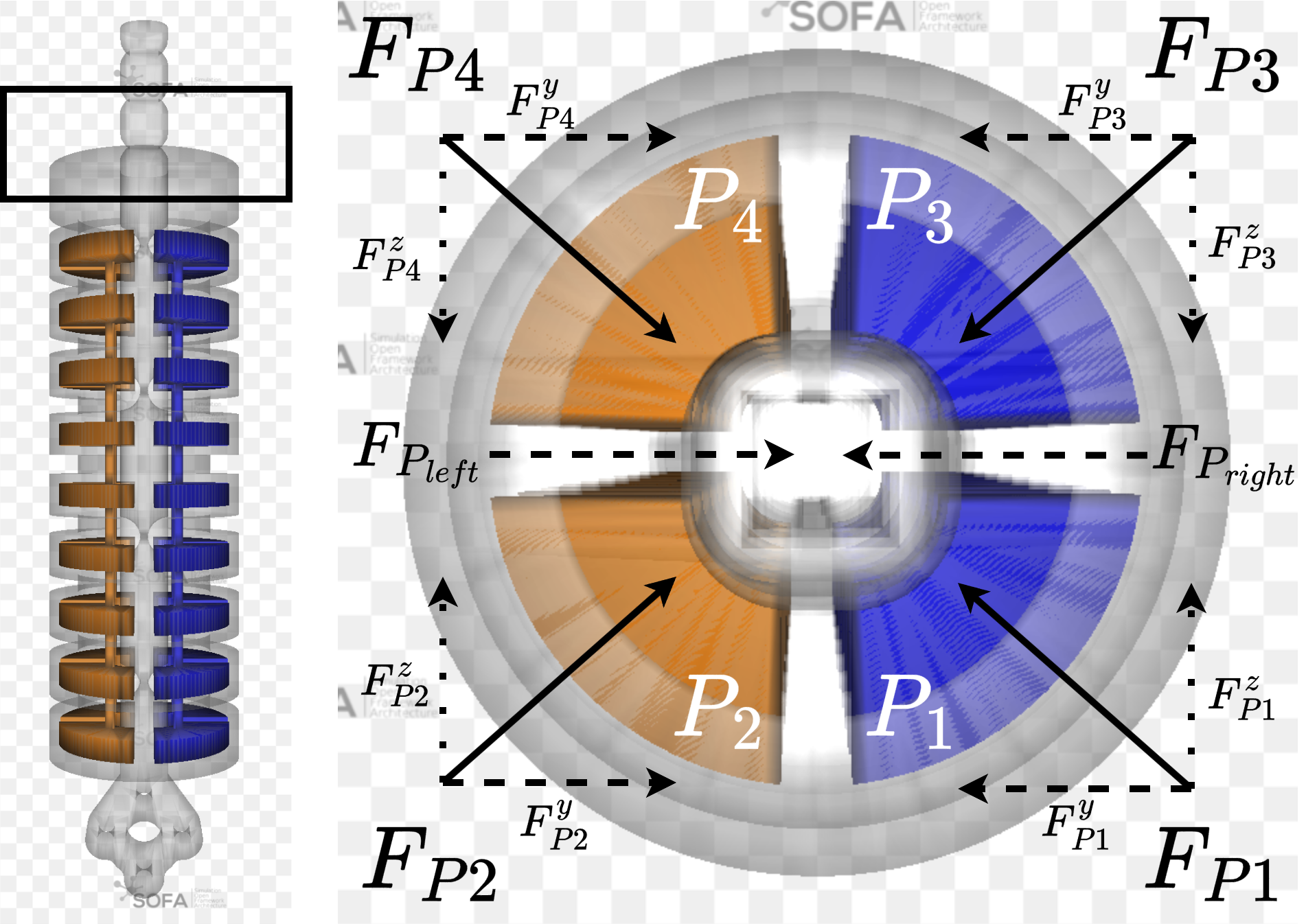}
    \caption{Experimental setup of the periodic control formulation. The arm cavities are merged on each side to produce horizontal forces during the differential pressure oscillation.} \vspace{-2mm}
    \label{fig:figure10}
\end{figure}

\noindent In this passive actuation formulation, the four cavities are merged into left and right pairs. As a result, the SPA bending is controlled by two independent pressure inputs $P_{left}$ and $P_{right}$, as in Eq. \ref{Eq02} and Eq. \ref{Eq03}.  This pneumatic control of the cavities results in a force towards the center of the arm that produces horizontal actuation as seen in Figure~\ref{fig:figure3}(c). The experimental setup is shown in Figure~\ref{fig:figure10}, where the effective forces $F_{P_{left}}$ and $F_{P_{right}}$ from the individual cavity force combination translates into horizontal displacement. \vspace{-1mm}

\begin{equation}\label{Eq02}
P_4=P_2; \quad P_{left}=P_4+P_2\ 
\end{equation} \vspace{-8mm}

\begin{equation}\label{Eq03}
P_3=P_1; \quad P_{right}=P_3+P_1
\end{equation} \vspace{-6mm}

Specifically, two time-dependent pressure signals, $P_{left}(t)$ and $P_{right}(t)$, are defined as sinusoidal signals with identical amplitude $A= 0.65Pa$ and frequency $f= 0.05 \text{Hz}$, centered around a constant offset pressure $P_0 = 0.65Pa$, as in Eq.~\ref{Eq2} and Eq.~\ref{Eq3}. The two signals are phase-shifted by $\pi$ radians such that they are symmetric about the mean value.
\vspace{-1mm}

\begin{equation}\label{Eq2}
P_{left}(t) = P_0 + A \sin(2\pi f t)
\end{equation} \vspace{-8mm}

\begin{equation}\label{Eq3}
P_{right}(t) = P_0 - A \sin(2\pi f t)
\end{equation} \vspace{-4mm}

\begin{figure}[h!]
    \centering
    \includegraphics[width=\linewidth]{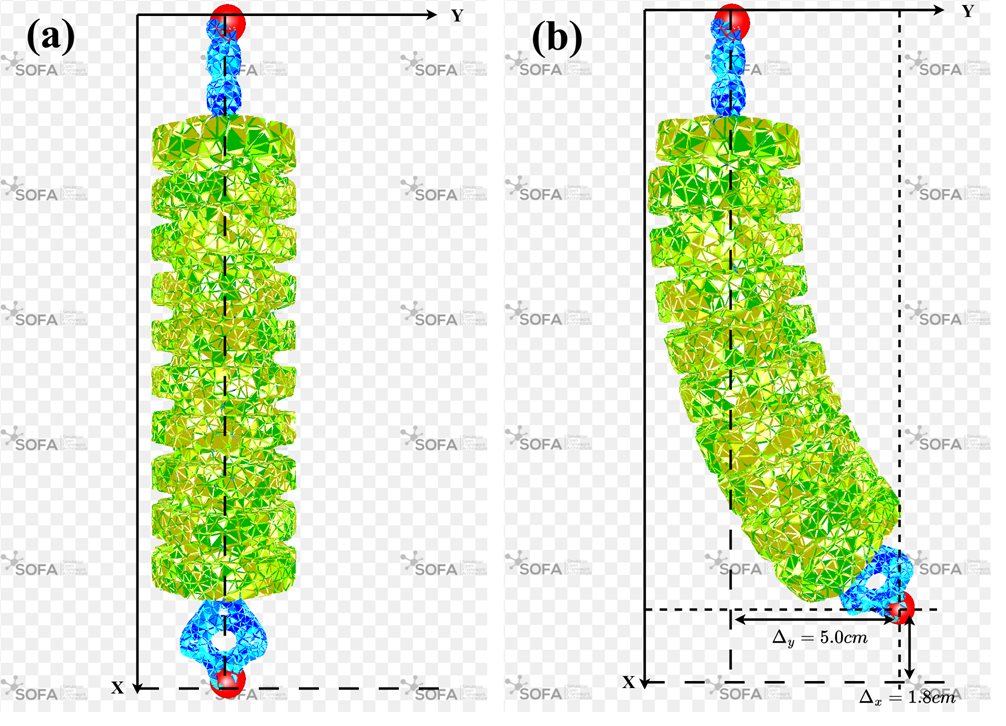}
    \caption{Pneumatic actuation effect under the passive control using differential sinusoidal pressure changes. The internal forces of the soft body are modeled by discretizing the finger and arm objects onto tetrahedral meshes following elastic and hyper elastic material laws.} \vspace{-4mm}
    \label{fig:figure4}
\end{figure} 

\begin{figure}[h!]
    \centering
    \includegraphics[width=\linewidth]{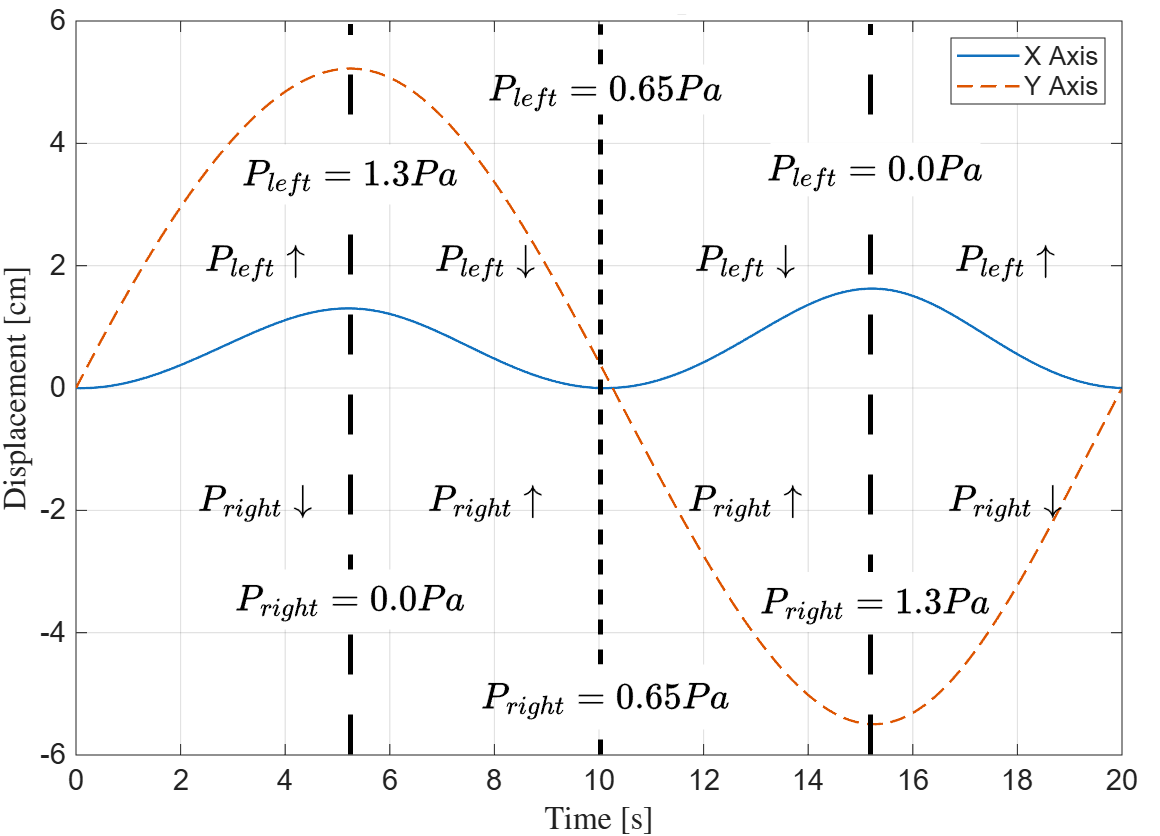}
    \caption{SPA position displacement during sinusoidal differential pressure changes on left and right cavities. Both cavities start at $0.65Pa$ and oscillate between $0Pa$ and $1.3Pa$. The deflection is mostly noticeable on the $XY$ plane with particularly high horizontal displacement on the $Y$ axis.} \vspace{-4mm}
    \label{fig:figure5}
\end{figure}

Figures~\ref{fig:figure4} (a) and (b) display respectively the initial unactuated state and the effect of the two-side cavity sinusoidal actuation during highest $P_{left}$ excitation on the physical model of the soft arm. The tetrahedral mesh displacement on $X$ and $Y$ axis is displayed in Figure~\ref{fig:figure5}. $P_{left}$ and $P_{right}$ initialize their pressures at $P_0$ and follow respectively an inverse sinusoidal signal, while the pressure inside the cavities oscillates in $[0, P_0 + A]$. The results gained, show a maximum bending displacement of $\Delta_Y = 5cm$ and $\Delta_X=1.8cm$ respectively. This formulation enables a three-state pneumatic configuration using a periodic signal for in-flight horizontal displacement of the soft arms.

\subsection{Error-based Pneumatic Actuation}

\begin{figure}[!t]
    \centering
    \includegraphics[width=\linewidth]{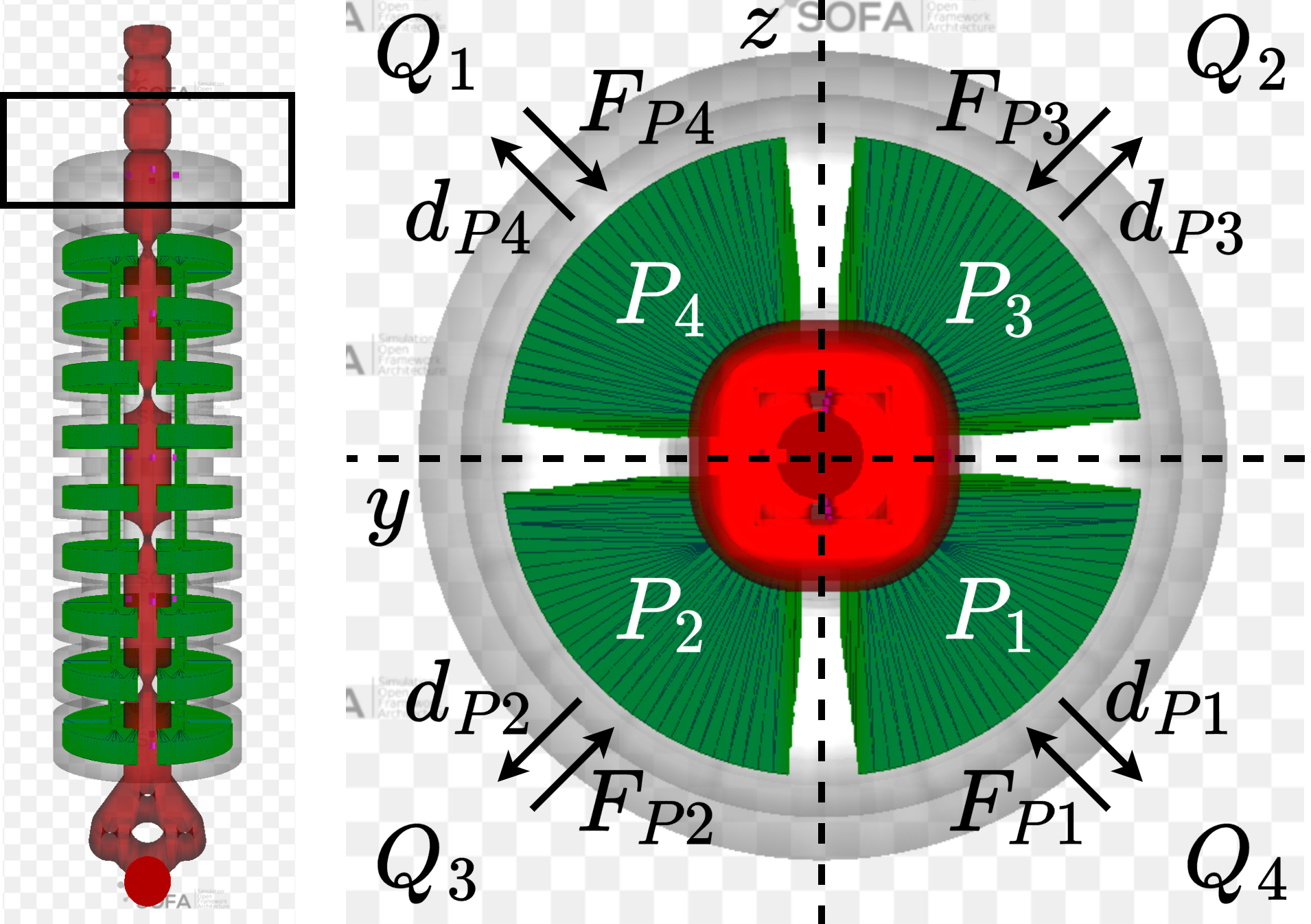}
    \caption{Experimental setup of the pressure controller formulation. The arm is divided into four quadrants with the main pressure cavity governing the deflection towards the quadrant on the diagonal opposite side.} \vspace{-5mm}
    \label{fig:figure9}
\end{figure}

\noindent This pressure controller formulation aims at regulating the individual cavity pressure with a classical positioning control approach. Let the current tip position of the soft actuator be $p$ and the desired value position be $p^\ast$. The Cartesian tracking error is defined as in Eq. \ref{error}.

\begin{equation}
    \label{error}
e = p^\ast - p
=
\begin{bmatrix}
e_x \\ e_y \\ e_z
\end{bmatrix}
=
\begin{bmatrix}
x^\ast - x \\
y^\ast - y \\
z^\ast - z
\end{bmatrix}
\end{equation}

\noindent where, $p^\ast$ = $[x^\ast, y^\ast, z^\ast]^{T}$ and $p$ = $[x, y, z]^{T}$. The actuator contains four pressure cavities arranged in a square configuration. The $y$ and $z$ components generate bending, while the $x$ component produces a common-mode (axial) deformation affecting all cavities equally. Figure~\ref{fig:figure9} shows this experimental description. 

The cavity demand vector $\mathbf{d} \in \mathbb{R}^4$, as defined in Eq.~\ref{dp_eqn}, is associated respectively with the opposite quadrant according to the relationship between the position error and the necessary effective force. Figure~\ref{fig:figure9} displays such relationship for each pressure cavity.

\[
d =
\begin{bmatrix}
d_{P1} \\
d_{P2} \\
d_{P3} \\
d_{P4}
\end{bmatrix}
=
Ae; \quad A =
\begin{bmatrix}
1 & -1 & -1 \\
1 & 1 & -1 \\
1 & -1 & 1 \\
1 & 1 & 1
\end{bmatrix}
\]

\begin{equation}\label{dp_eqn}
    \begin{matrix}
d_{P1} = e_x - e_y - e_z; \quad d_{P2} = \;\;e_x + e_y - e_z \\
d_{P3} = e_x - e_y + e_z; \quad d_{P4} = \;\;e_x + e_y + e_z
\end{matrix}
\end{equation}

\noindent Unnecessary actuation due to small errors is prevented with a deadband $\delta=0.05$ on each cavity demand as in Eq.~\ref{eq6}.

\begin{equation}\label{eq6}
\tilde d_i =
\begin{cases}
0, & |d_i| < \delta, \\
d_i, & \text{otherwise}.
\end{cases}    
\end{equation}

\noindent A Proportional Integral (PI) controller is designed to control the  magnitude of the cavity demand vector. The scalar control error $e(t)$ is defined in Eq.~\ref{eq7} as the Euclidean norm of the cavity demands. The PI control signal $u(t)$, with parameters $k_p=2 \times 10^{-6}; \  k_i=2 \times 10^{-8}$ is described as in Eq.~\ref{eq8}. \vspace{-2mm}

\begin{equation}\label{eq7}
    e(t) = \|\tilde{d}(t)\|_2
= \sqrt{\sum_{i=1}^{4} \tilde d_i^2}
\end{equation} \vspace{-4mm}

\begin{equation}\label{eq8}
    u(t) =
k_p e(t)
+
k_i \int_0^t e(\tau)\, d\tau
\end{equation}

\begin{figure}[!h]
    \centering
    \includegraphics[width=\linewidth]{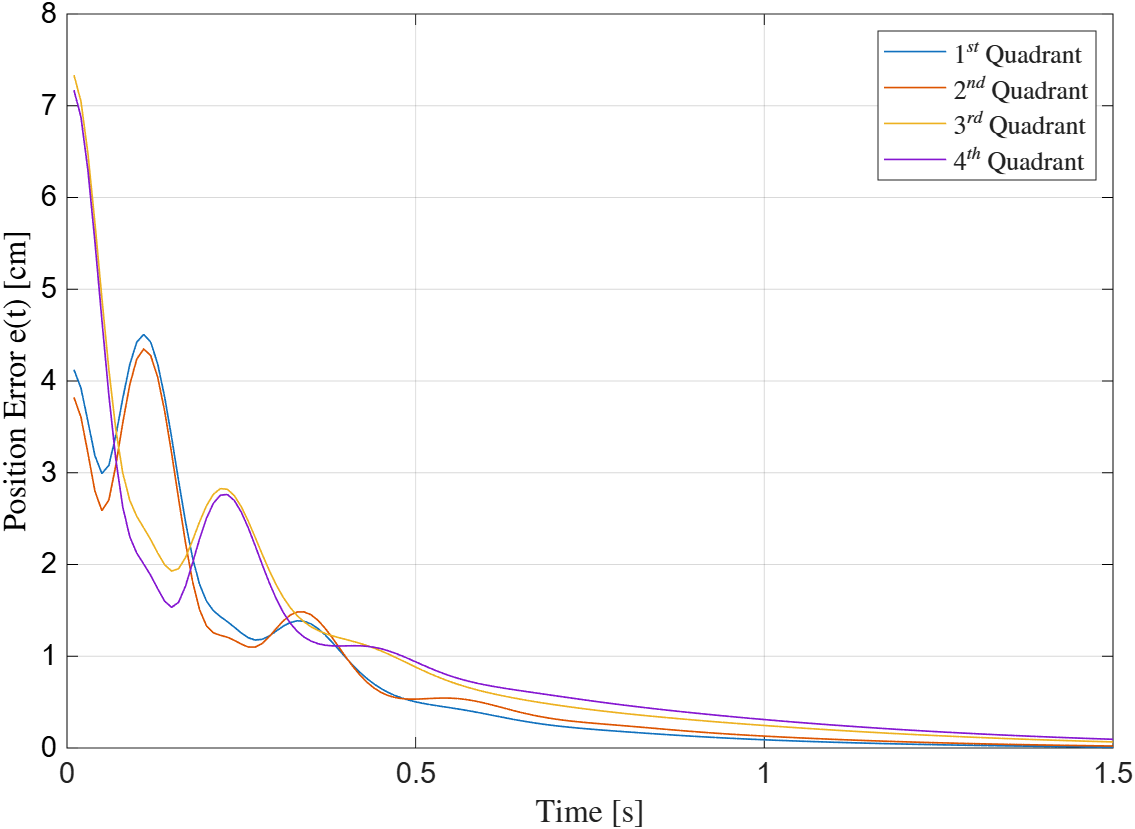}
    \caption{Scalar error signal $e(t)$ from position controller during the four quadrant experiments. The pneumatic arm's pressure is regulated based on the distance to a point selected from each quadrant.} \vspace{-2mm}
    \label{fig:figure7}
\end{figure}

\noindent In discrete time with sampling interval $\Delta t$, the error and control signals are redefined as in Eq.~\ref{eq9}, respectively.   \vspace{-1mm}

\begin{equation}\label{eq9}
\begin{matrix}
e_k = \|\tilde{d}_k\|_2; \quad I_k = I_{k-1} + e_k \Delta t \\ \\
u_k = k_p e_k + k_i I_k
\end{matrix}    
\end{equation}

\noindent where $k$ represents the discretization step given by $\Delta t$. Figure~\ref{fig:figure7} shows the scalar error signal over time, while achieving the target position on each quadrant $p^\ast = (0, \pm1.5, \pm1.0) (cm)$. Results show that with the given computed control, the soft arm deflects accurately towards the desired position in each quadrant, while the error becomes zero in a finite time.

The total pressure increment $u_k$ is distributed proportionally to the signed cavity demands. Defining the $L^1$ normalization factor $\alpha_k$, the pressure increment $\Delta P_{k,i}$ for cavity $i$ is defined as in Eq. \ref{eq10}. \vspace{-1mm} 

\begin{equation}\label{eq10}
\alpha_k = \sum_{i=1}^{4} |\tilde d_{k,i}|;  \quad 
\Delta P_{k,i} =
\begin{cases}
u_k \dfrac{\tilde d_{k,i}}{\alpha_k}, & \alpha_k > 0, \\
0, & \alpha_k = 0
\end{cases}    
\end{equation}

\begin{figure*}[!t]
    \centering
    \includegraphics[width=\linewidth]{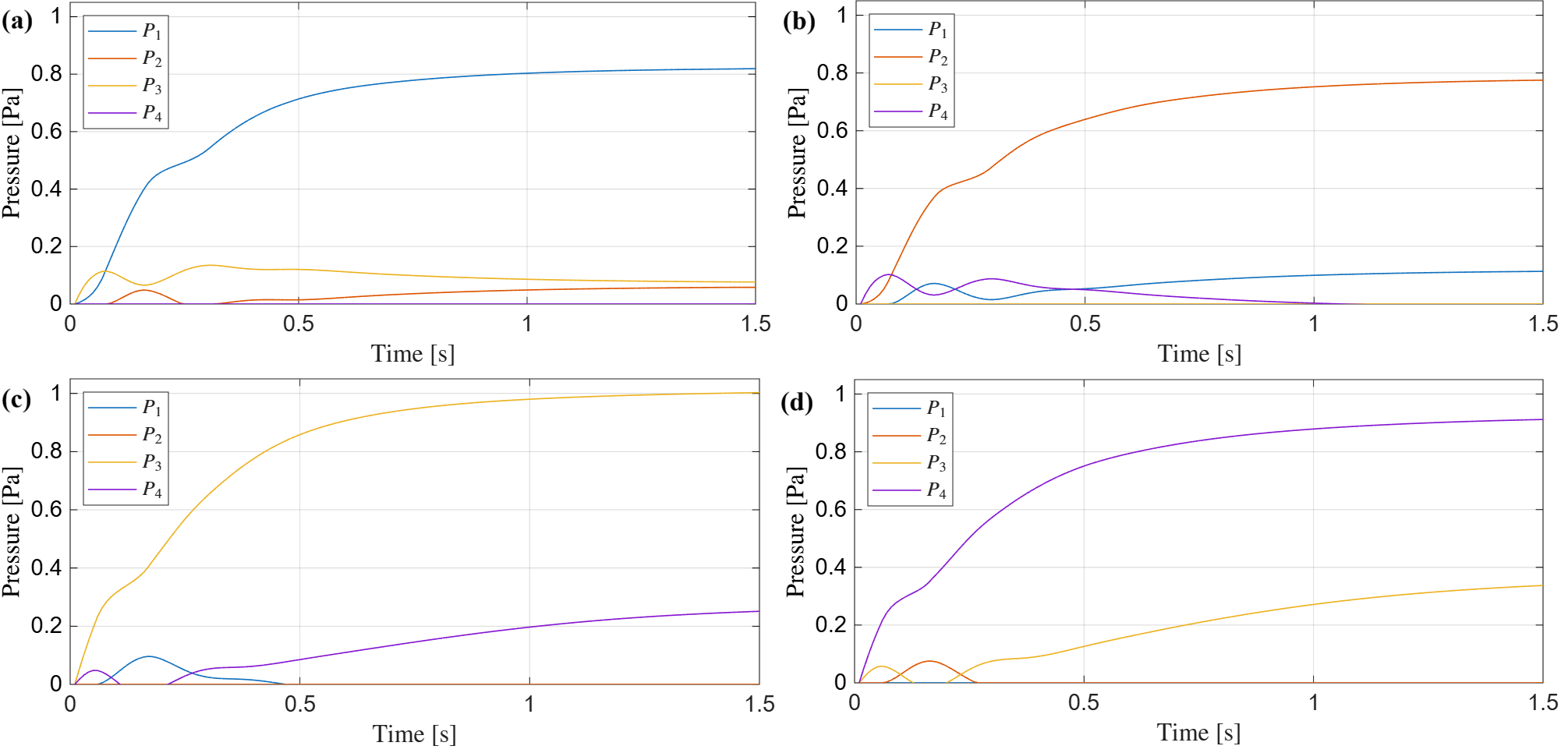}
    \caption{Distributed control signal i.e. individual cavity pressure for every quadrant experiment $Q_{1-4} \rightarrow$ (a)-(d). A point $p^\ast = (0, \pm1.5, \pm1.0) (cm)$ is selected as target position during the experiment. The control algorithm regulates the pressure according to the error signal and distributes the pressure over the four cavities based on the effective actuation magnitude.} \vspace{-4mm}
    \label{fig:figure6}
\end{figure*}

\noindent Let $P_k$ denote the vector of pneumatic actuation in all four cavities. The pressure variable takes continuous values within the physical bounds. A clipping function is applied elementwise to restrict the control signal between the pneumatic actuation boundaries $P_{min}=0.0Pa; \ P_{max}=1.3Pa$. \vspace{-2mm}

\begin{equation}\label{eq11}
    \mathrm{clip}(x,a,b) = \min(\max(x,a), b)
\end{equation} \vspace{-6mm}

\begin{equation}\label{eq12}
P_{k+1}
=
\mathrm{clip}
\left(
P_k + \Delta P_k,
\; P_{\min},
\; P_{\max}
\right)    
\end{equation}

\noindent where $P_k = [P_{k,1}, \ P_{k,2}, \ P_{k,3}, \ P_{k,4}]^T$. Figure \ref{fig:figure6} (a)-(d) respectively shows the distributed control signal evolution $P_k$ during each of the quadrant experiments $Q_{1-4}$. For example, to achieve a target point in quadrant $Q_{1}$, the computed control signal pneumatically actuates $P_1$ to vary from $0$ to $0.8Pa$, while the actuation to all the other three cavities remains minimal. Figures \ref{fig:figure6} (c) and (d) show a varied response wherein to achieve the desired target in $Q_{3}$ and $Q_{4}$, the control signal is distributed among two cavities with a net increase of approximately $0.2Pa$. 

In the unactuated state, the soft arm is subject to gravitational forces causing a natural deflection towards quadrants $Q_{3}$ and $Q_{4}$. When the target is below the resting arm position, the bending limits of the elastic material are reached and an increased pressure is needed to achieve correct positioning, as seen in Figure~\ref{fig:figure6} (c) and (d). This formulation enables coordinated 3D motion control using a single PI regulator while preserving directional bending behavior.

\section{Conclusion and future works} 
\noindent In this work, the SOFA framework is used to develop a dynamic finite-element simulation model of a pneumatic morphing soft quadrotor along with pneumatic control capabilities. The results show that the FEM-based model together with the proportional-integral controller can achieve any desired position in any of the four quadrants under the given pneumatic actuation constraint. Both periodic and error-based pneumatic actuation approaches demonstrate the control of the internal cavities under different configurations. Future work focuses on developing an adaptive controller for the soft morphing quadrotor and a reinforcement learning-based navigation policy, using the finite element model, with sim-to-real transfer (domain randomization and safety constraints) for deployment on the real soft-drone hardware.

\bibliographystyle{unsrt}
\bibliography{references.bib}
\end{document}